%% file: main.tex
\documentclass[runningheads]{llncs}
\usepackage{graphicx}
\usepackage{comment}
\usepackage{amsmath,amssymb} 
\usepackage{color}
\usepackage{epsfig}
\usepackage{paralist, tabularx}
\usepackage{flushend}
\usepackage{booktabs}
\usepackage{ulem}
\usepackage{subcaption}

\newcommand{\ra}[1]{\renewcommand{\arraystretch}{#1}}

\usepackage[pagebackref=true,breaklinks=true,colorlinks,bookmarks=false]{hyperref}

\usepackage{microtype}

\input{macro.tex}

\begin{document}
\pagestyle{headings}
\mainmatter

\title{QuEST: Quantized Embedding Space\\for Transferring Knowledge} 

\author{Himalaya Jain\inst{1} \and Spyros Gidaris\inst{1} \and Nikos Komodakis\inst{2,3} \and Patrick P{\'e}rez\inst{1} \and Matthieu Cord\inst{1,4}}

\authorrunning{H. Jain et al.}

\institute{\textsuperscript{1}Valeo.ai \enspace \textsuperscript{2}University of Crete \enspace \textsuperscript{3}LIGM \enspace \textsuperscript{4}Sorbonne University}
\maketitle

\begin{abstract}
Knowledge distillation refers to the process of training a student network to achieve better accuracy by learning from a pre-trained teacher network. 
Most of the existing knowledge distillation methods direct the student to follow the teacher by matching the teacher's output, feature maps or their distribution.
In this work, we propose a novel way to achieve this goal: by distilling the knowledge through a \textit{quantized visual words} space.
According to our method, the teacher's feature maps are first quantized to represent the main visual concepts (i.e., visual words) encompassed in these maps and then the student is asked to predict those visual word representations.
Despite its simplicity, we show that our approach is able to yield results that improve the state of the art on knowledge distillation for model compression and transfer learning scenarios. 
To that end, we provide an extensive evaluation across several network architectures and most commonly used benchmark datasets.
\end{abstract}

\section{Introduction}

\input{intro.tex}
\section{Related work} \label{sec:related_work_section}
\input{related.tex}

\section{Approach} \label{sec:approach_section}
\input{approach.tex}
\section{Experiments} \label{sec:experiments_section}
\input{experiments.tex}
\section{Conclusions} \label{sec:conclusion_section}

Our work deals with the important learning problem of knowledge distillation.
The goal of knowledge distillation is to improve the accuracy of a student network by exploiting the learned knowledge of a pre-trained teacher network. 
To that end, we follow the common paradigm of transferring the knowledge encoded on the learned teacher features. 
However, instead of performing the distillation task in the context of the initial feature space of the teacher network, we transform it to a new quantized space.

Specifically, our distillation method first densely quantizes the teacher feature maps into visual words and then trains the student to predict this quantization based on its own feature maps. 
By solving this task the student is forced to align its feature maps with those of the teacher network while ignoring unimportant ``feature" details, thus facilitating efficient knowledge transfer between the two networks. 
To demonstrate the effectiveness of our distillation method,
we exhaustively evaluate it on two very common knowledge distillation scenarios,
model compression (i.e., on ImageNet, CIFAR-100 and CIFAR-10 datasets) and transfer learning to small-sized datasets (i.e., ImageNet to MIT-67 datasets), 
and across a variety of deep network architectures. 
Despite its simplicity, our method manages to surpass prior work, achieving new state-of-the-art results on a variety of benchmarks.

\bibliographystyle{splncs04}
\bibliography{egbib}

\appendix
\input{appendix}

\end{document}

%% file: macro.tex
\usepackage{amsmath}
\usepackage{amssymb}
\usepackage{dsfont}
\usepackage{bm}
\usepackage{xcolor,soul,colortbl}
\usepackage{array}
\usepackage{multirow}
\usepackage{dashrule}
\usepackage{breqn}
\usepackage{mathtools}

\newcommand{\R}[0]{\mathbb R}
\newcommand{\mv}[1]{\ensuremath{\bm{#1}}} 
\newcommand{\mm}[1]{\ensuremath{\bm{#1}}} 

\newcommand{\s}[1]{#1_S} 
\newcommand{\T}[1]{#1_T} 

\newcommand{\bz}{\mv z}

\makeatletter
\DeclareRobustCommand\onedot{\futurelet\@let@token\@onedot}
\def\@onedot{\ifx\@let@token.\else.\null\fi\xspace}
\def\ie{i.e.}

\def\etal{et al.}
\makeatother

%% file: intro.tex
Knowledge distillation is an interesting learning problem with many practical applications.
It was initially introduced by Hinton \etal~\cite{hinton2015distilling} (KD method) as a means to achieve model compression.
The main idea is to train a network using the output of another pre-trained network.
Specifically, KD trains a low capacity ``student network" (i.e., compressed model) to mimic the \textit{softened classification predictions} of a higher capacity pre-trained ``teacher network" (i.e., original uncompressed model).
Adding such an auxiliary objective to the standard training loss of the student network, leads to learning a more accurate model.
Apart from model compression, knowledge distillation has also been shown to be beneficial to semi-supervised learning~\cite{laine2016temporal,tarvainen2017mean} and transfer learning~\cite{li2017learning} problems.

\begin{figure*}[t!]
    \centering
    \begin{tabular}{ccc}
    \includegraphics[width=0.33\linewidth]{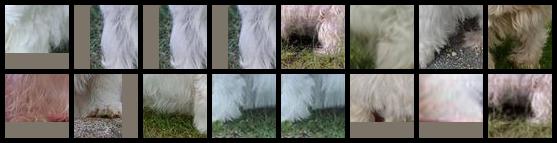} & \includegraphics[width=0.33\linewidth]{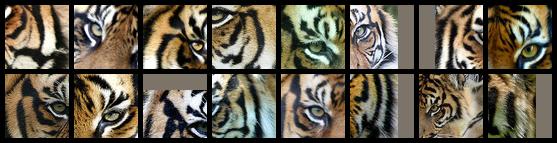} & \includegraphics[width=0.33\linewidth]{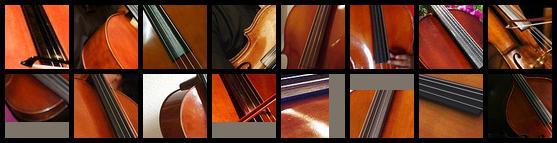}\\
    \includegraphics[width=0.33\linewidth]{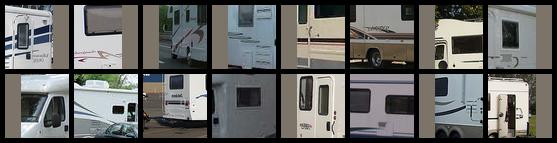} & \includegraphics[width=0.33\linewidth]{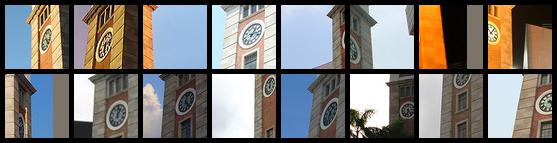} & \includegraphics[width=0.33\linewidth]{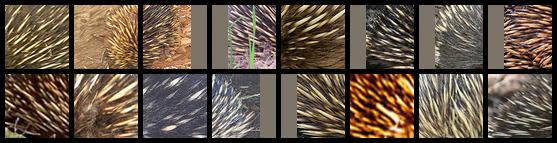}\\    
    \end{tabular}
    \caption{
    \textbf{Teacher-word clusters from layer4 of ResNet34.} 
    For each cluster we depict the 16 patch members with the smallest Euclidean distance to the cluster centroid.
    We see that they encode localized mid-to-high level image patterns
    }
    \label{fig:visual_words}
\end{figure*}

Since KD, several other methods have been proposed for knowledge distillation 
\cite{cho2019EKD_iccv,furlanello2018born,heo2019knowledge,kim2018paraphrasing,liu2019_cvpr,passalis2018learning,romero2014fitnets,yang2018knowledge,zagoruyko2017AT}. 
A popular paradigm is that of 
transferring knowledge through intermediate features of the two networks~\cite{romero2014fitnets,zagoruyko2017AT}.
For instance, in FitNet~\cite{romero2014fitnets} the student is trained to regress the raw feature maps of the teacher.
The main intuition is that 
these features tend to encode useful semantic knowledge for transfer learning.
However, empirical evidence~\cite{ahn2019variational,heo2019comprehensive,heo2019knowledge,yim2017gift,zagoruyko2017AT} suggests that naively regressing the teacher features might be a 
difficult task which over-regularizes the student, thus leading to a suboptimal knowledge transfer solution.
This can be because \textbf{(a)}
the exact feature magnitudes (i.e., feature ``details") of the feature maps are not semantically important aspects of the teacher network in the context of knowledge distillation, and/or \textbf{(b)} features behaviour largely depends on the architecture design (e.g., ResNet versus MobileNet) and capacity (i.e., width and depth) of a network.
For instance, 
AT~\cite{zagoruyko2017AT} shows that transferring attention maps (feature maps aggregated into a single channel) might be a better option in many cases; 
FSP~\cite{yim2017gift} opts to remove the spatial information by considering only the correlations among feature channels;
AB~\cite{heo2019knowledge} transfers binarized versions of feature maps;
and VID~\cite{ahn2019variational} proposes a regression loss (inspired from information theory) that is meant to ``ignore" difficult to regress channels.

In our work we propose to address the knowledge transfer task in the context of \textit{a spatially dense quantization of feature maps into visual codes}.
Specifically, first we learn with k-means a vocabulary of deep features from the teacher network, which we call visual teacher-words.
Then, given an image, we use this vocabulary to quantize/assign each location of the teacher's feature maps into the closest visual words. 
As distillation task, the student has to predict these visual teacher-word assignments maps using an auxiliary convolutional module.
Hence, our method departs from prior work and essentially \textit{transfers discretized (into words) versions of the feature maps}.
Besides being remarkably simple, this strategy
has the advantage of aligning the student's behavior with that of the teacher by focusing only on the main local visual concepts (\ie, teacher-words) that the teacher has learnt to detect in an image (see Figure \ref{fig:visual_words}).
As a result, it ignores unimportant feature ``details" that are difficult to regress, thus avoiding over-regularizing the student.
Compared to prior work (e.g.,~\cite{heo2019knowledge,yim2017gift,zagoruyko2017AT}),
we argue that our strategy of transferring visual word representations is more effective as it does not withhold important aspects/cues of the teacher's feature maps.

We extensively evaluate our method 
on two knowledge distillation scenarios (i.e., model compression and transfer learning on small-sized datasets),
with various pairs of teacher-student architectures,
and across a variety of datasets. 
In all cases our simple but effective approach achieves state-of-the-art results.

%% file: related.tex
The aim of knowledge distillation is to transfer knowledge from a trained teacher network to a student network. 
To carry out the knowledge transfer, the student is trained to imitate some facets of the teacher network. 
For instance, 
Buciluǎ \etal~\cite{bucilua2006model} proposed to train the student as an approximator of a large ensemble of teacher networks by predicting the output of this network ensemble and
Hinton \etal~\cite{hinton2015distilling} proposed 
to train the student to output the same softened predictions (i.e., classification probabilities) as the trained teacher network.
   
The teacher and student are deep networks, having a sequential structure that is known to learn hierarchical representations of increasing abstraction.
Inspired by this, some methods propose to match the intermediate activation or feature maps of the teacher as the distillation task for the student. This encourages the student to follow the intermediate solutions produced by the teacher. FitNet \cite{romero2014fitnets} proposes to train the student with an additional layer to regress the feature maps of the teacher. While AT \cite{zagoruyko2017AT} builds ``attention maps" by aggregating feature maps, and $\ell_2$ error between normalized attention maps of student and teacher is used as distillation loss. In FSP~\cite{yim2017gift}, the difference between Gram matrices of feature maps of the two networks is minimized for distillation.

Another line of works  
focuses on matching distributions of feature maps rather than feature maps themselves. In NST \cite{huang2017like}, the maximum mean discrepancy between the distributions of feature maps of teacher and student is minimized.  
VID \cite{ahn2019variational} and
CRD \cite{tian2019contrastive} consider maximizing mutual information between the two networks as the knowledge distillation task. 
Mutual information is maximized by maximizing variational lower bound (VID) or contrastive loss (CRD). 

SP \cite{tung2019SP_iccv} departs from matching feature maps or their distributions. 
This relieves the student network from the demanding task of copying the teacher network, which could be too ambitious given the difference in their capacity and architecture.
SP proposes instead a similarity preserving constraint by using difference in pairwise similarity, computed on a mini-batch, as distillation loss. 
Similar methods are proposed in other recent works RKD \cite{park2019RKD_cvpr} and CC \cite{peng2019CC_iccv}.

In our work, similar to SP, we propose a different space for distillation than feature space. 
We propose to use a quantized space where the teacher's feature maps are 
encoded by quantization with learned visual teacher-words.
These visual words are learned by k-means clustering on the feature maps of the teacher thus, they represent useful semantic concepts. 
With the proposed quantized space we concentrate more on the important semantic concepts and their spatial correlation for knowledge distillation.

We note that quantizing features into visual words is key a ingredient of 
the bag-of-visual-word techniques that were extensively used in the past~\cite{csurka2004visual,jegou2010aggregating,perronnin2007fisher,sivic2006video,tolias2013aggregate}.
Although these visual-word techniques have now been used with deep learning~\cite{arandjelovic2016netvlad,gidaris2020learning,girdhar2017actionvlad}, our work is the first to leverage a visual-word strategy with deep learning for knowledge distillation.

%% file: approach.tex
\begin{figure*}[t!]
    \centering
    \includegraphics[width=0.85\textwidth]{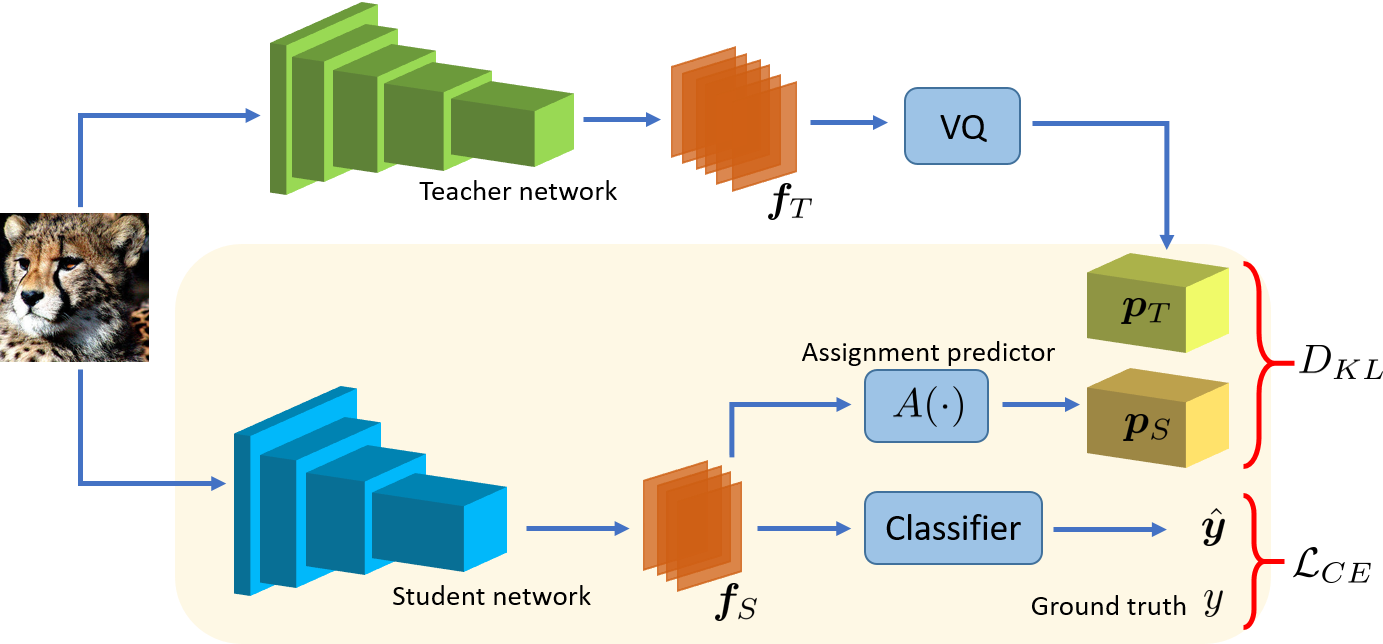}
    \caption{
    Overview of the proposed method for knowledge transfer based on visual word assignments. 
    The feature maps $\T{\mm{f}}$ of the teacher are converted into a visual words representation $\T{\mv p}$ via vector quantization (VQ) with soft-assignment.
    Then, as distillation task, the student has to predict the teacher's words assignments $\T{\mv p}$ using its own feature maps $\s{\mm{f}}$ and an auxiliary assignment prediction module $A$.
    The distillation loss is the KL-divergence $D_{KL}$ of the predicted word assignments $\s{\mv p}$ with $\T{\mv p}$.
    The visual word vocabulary that is used in VQ is learned off-line by applying the k-means clustering algorithm to teacher feature maps extracted from the training dataset
    }
    \label{fig:approach}
\end{figure*}

\subsection{Preliminaries}

Here we briefly explain the learning setting of knowledge distillation methods.
Let $S$ be a student network that we want to train using a dataset of $(\mv x, y)$ examples, where $\mv x$ is the image and $y$ is the label.
In the standard supervised learning setting the teacher $S$ is trained to minimise for each example $(\mv x, y)$ the classification loss 
$\mathcal{L}_{\textit{CLS}}(S) = \mathcal{L}_{\textit{CE}}(y, \sigma(\s\bz))$, 
where $\mathcal{L}_{\textit{CE}}$ is the cross-entropy loss,
$\s\bz = S(\mv x)$ are the classification logits predicted from $S$ for the image $\mv x$, and $\sigma(\cdot)$ is the softmax function.
Knowledge distillation methods assume that there is available a pre-trained teacher network $T$, which has higher capacity than that of the student $S$.
The goal is to exploit the pre-trained knowledge of teacher $T$ for training a better (i.e., more accurate) student $S$ than the standard supervised learning setting.
To that end, they define for each training example $(\mv x, y)$ an additional loss term $\mathcal{L}_{\textit{DIST}}(S; T)$ based on the pre-trained teacher $T$. 
For instance, in the seminal work of Hinton~\etal\cite{hinton2015distilling}, 
the additional distillation loss is the cross entropy of the two network output softmax distributions for the same training example $(\mv x, y)$:
\begin{equation}
\begin{split}
\mathcal{L}_{\textit{DIST}}(S; T) = \rho^2 \mathcal{L}_{\textit{CE}} \Big(\sigma(\frac{\T\bz}{\rho}), \sigma(\frac{\s\bz}{\rho})\Big),
\end{split}
\label{eq:ldist}
\end{equation}
where $\T\bz = T(\mv x)$ are the logits produced from $T$ for the image $\mv x$, and $\rho>0$ is a temperature that softens (i.e., lowers the peakiness) of the two softmax distributions.
Through this loss, the student learns from the teacher by mimicking the teacher's output on the training data.
Therefore, the final objective that the student $S$ has to minimize per training example $(\mv x, y)$ is:
\begin{equation}
\mathcal{L} = \alpha \mathcal{L}_{\textit{CLS}}(S) + \beta \mathcal{L}_{\textit{DIST}}(S; T),
\label{eq:ltotal}
\end{equation}
where $\alpha$ and $\beta$ are the weights of the two loss terms.

In the next subsection we explain the distillation loss $\mathcal{L}_{\textit{DIST}}(S; T)$ that we propose in our work.

\subsection{Distilling visual teacher-word assignments} \label{sec:distillation_task}

An overview of our approach is provided in Figure~\ref{fig:approach}.
The proposed distillation task first converts the teacher feature maps into visual word assignments
and then trains the student to predict those assignments from its own feature maps.

\subsubsection{Converting the teacher feature maps into visual word assignments.} \label{sec:teacher_quantization}

Given an image $\mv x$, let $\T{\mm{f}}\in\R^{\T C\times\T H\times\T W}$ be the feature map (with $\T C$ channels and $\T H\times\T W$ spatial dimensions) that $T$ generates at one of its hidden layers, and $\T{\mm{f}^{(h, w)}} \in \R^{\T C}$ be the feature vector at the location $(h, w)$ of this feature map. 
We quantize $\T{\mm{f}}$ in a spatially dense way using a predefined vocabulary $V = \{\mv v_k \}_{k=1\dots K}$ of $K$ $\T C$-dimensional visual teacher-word embeddings.
Specifically, we first compute for each location $(w,h)$ the squared Euclidean distances between the feature vector $\T{\mm{f}^{(h, w)}}$ and the $K$ teacher-words:
\begin{equation}
    \mv d^{(h, w)} = \Big[\left\lVert \mv v_k -  \T{\mm{f}^{(h, w)}}\right\rVert^2_2\Big]_{k=1\dots K}.
\label{eq:distances}
\end{equation} 
Then, using $\mv d^{(h,w)}$ we compute 
the $K$-dimensional soft-assignment vector $\T{\mv p^{(h, w)}}$, which lies on a $K$-dimensional probability simplex, as:
\begin{equation}
    \T{\mv p^{(h, w)}} = \sigma(-\mv d^{(h, w)}/\tau), 
\label{eq:softquant}    
\end{equation}
where $\tau>0$ is a temperature value used for controlling the softness of the assignment. 
The visual teacher-word vocabulary $V$ is learned off-line by k-means clustering feature vectors $\T{\mm{f}^{(h, w)}}$ extracted from the available training dataset. 
As we see in Figure~\ref{fig:visual_words}, the vocabulary $V$ includes the main local image patterns that are captured by the teacher's feature maps (e.g., a dog's leg on grass, the eye of a tiger, or a clock tower).
Hence, $\T{\mv p^{(h, w)}}$ essentially encodes what semantic concepts (defined by $V$) the teacher network detects at location $(h,w)$ of its feature map.

\paragraph{Soft-assignment vs. hard-assignment.}
In our experiments we noticed that the optimal $\tau$ value produces very peaky soft-assignments, e.g., the softmax probability for the closest visual teacher-words is on average around $0.996$, which is very close to hard assignment.
However, very peaky soft-assignment still leads to slightly better students, 
which can be attributed to either 
\textbf{(a)} the fact that, as in KD~\cite{hinton2015distilling}, the probability mass for the remaining visual teacher-words carries useful teacher knowledge, or
\textbf{(b)} soft-assignments regularize the visual teacher-words prediction task that the student has to perform~\cite{yuan2019revisit}.

\subsubsection{Predicting the teacher's visual word assignments.} \label{sec:predict_quantizations}

The distillation task that we propose is to train the student $S$ to predict the teacher soft-assignment map $\T{\mv p}$ 
(of visual words) based on its own feature map $\s{\mm{f}}\in\R^{\s C\times\s H\times\s W}$.
As $\T{\mv p}$ encodes semantic information with spatial structure, we posit that, to predict $\T{\mv p}$ the student must build an understanding of semantics similar to the teacher.
We assume for simplicity that the student's feature map $\s{\mm{f}}$ has the same spatial size as that of the teacher\footnote{
Otherwise, we simply down-sample the biggest feature map to the size of the smaller one with an (adaptive) average pooling layer.}, i.e., $\s H = \T H$ and $\s W = \T W$.
We also note that we impose \textit{no constraint on the number of student feature channels} $\s C$.
Therefore, to predict $\T{\mv p}$, we use an \textit{assignment predictor} $A$, which consists of one cosine-similarity-based convolutional layer with $1\times1$ kernel size followed by a learnable scaling factor and a softmax function. 
Specifically, at each location $(h,w)$, 
the student predicts the $K$-dimensional soft-assignment vector $\s{\mv p^{(h, w)}}$ from its feature vector $\s{\mm{f}^{(h, w)}}$, as:
\begin{equation}
    \s{\mv p^{(h, w)}} = A(\s{\mv f}^{(h,w)}) = \sigma\Bigg(\Bigg[\gamma\frac{\langle \mm W_i, \s{\mv f}^{(h,w)} \rangle}{\lVert\mm W_i\rVert\lVert\s{\mv f}^{(h,w)}\rVert}\Bigg]_{i=1\dots K}\Bigg),
\label{eq:apredictor}
\end{equation}
where $\gamma$ is the learnable scaling factor, $\mm W \in \R^{\s C \times K}$ are the parameters of the convolutional layer, and $\mm W_i$ is the $i$-th column of $\mm W$.
Hence, $\mm W$ plays essentially the role of a new student vocabulary, and soft assignment is done according to cosine similarity (instead of Euclidean distance).

Finally,
we define our distillation loss $\mathcal{L}_{\textit{DIST}}(S; T)$ (for a training example) as the summation over all the locations $(h, w)$ of the KL-divergence $D_{KL}$ of the predicted soft-assignment distribution $\s{\mv p^{(h, w)}}$ from the target distribution $\T{\mv p^{(h, w)}}$:
\begin{equation}
    \mathcal{L}_{\textit{DIST}}(S; T) = \sum_{h,w} D_{KL}(\T{\mv p^{(h, w)}} \| \s{\mv p^{(h, w)}})\enspace,
\label{eq:our_distil_loss}
\end{equation}
where $D_{KL}(\T{\mv p^{(h, w)}} \| \s{\mv p^{(h, w)}})$ consists of the cross-entropy of $\T{\mv p^{(h, w)}}$ and $\s{\mv p^{(h, w)}}$, and the entropy of $\T{\mv p^{(h, w)}}$. 
The latter is independent to the student and thus does not actually affect the training. 
We minimize the total loss $\mathcal{L} = \alpha \mathcal{L}_{\textit{CLS}}(S) + \beta \mathcal{L}_{\textit{DIST}}(S; T)$ of the student network over $W$, $\gamma$, and the parameters of $S$.
Note that it is possible to apply the proposed distillation loss to more than one feature levels.
In this case the final distillation loss is the sum of all the per-level losses.

\subsection{Discussion}\label{sec:discussion}

\textit{Transferring ``discretized'' representations.}
In our method, the teacher's feature maps are essentially discretized into visual words 
(i.e., as already described, the softmax probability for the closest visual teacher-words is on average around $0.996$).
Therefore, by minimizing the KL-divergence between the predicted visual word assignments $\s{\mv p}$ and the teacher assignments $\T{\mv p}$,
we essentially ``discretize'' the student feature maps according to the ``discretization" at the teacher side.
The difference is that in the student case, the visual vocabulary is implicitly defined by the parameters $W$ of the assignment predictor $A$, and the assignment is done via cosine similarity instead of Euclidean distance.\footnote{In Sec.\,\ref{sec:appendix_cs} of Appendix, we discuss the use of cosine similariy on student side.}
Hence, the student feature maps must learn to discriminate (i.e., to cluster together) the same local image patterns (represented by the visual words) as the teacher feature maps.
This way, the student learns to align its feature maps with those of the teacher network both spatially and semantically, but without being forced to regress the exact feature ``details'' of its teacher (which can over-regularize the student).

\paragraph{Posing feature transfer as a classification problem.}
In a sense, our method converts the direct feature regression problem into a classification one where the teacher-words are the classification prototypes.
As has been shown in prior works (e.g.,  see methods based on contrastive learning~\cite{oord2018representation,tian2019contrastive}), this leads to a training task that is easier to solve/optimize and, as a result, constitutes a far more effective learning strategy.

\paragraph{Relation to KD method~\cite{hinton2015distilling}.}
Our method relates to KD in the sense that it also tries to ``mimic'' softened classification predictions.
However, in our case those classification predictions are spatially dense and over the visual teacher-words,
which represent visual concepts that are not only more localized but also much more numerous than the available image classes. 
Due to this fact, our distillation loss leads to a more efficient (i.e., richer) knowledge transfer.
We should note at this point that many prior methods combine their proposed distillation loss with the KD loss.
In our case, however, we empirically observed that adding the KD loss to our method is not required as it does not offer any significant performance improvement, which further verifies the effectiveness of our approach.

%% file: experiments.tex
In this section, we experimentally evaluate our proposed approach and compare it with several state-of-the art knowledge distillation methods.
In the remainder of this section, we first compare our approach against prior art with extensive experiments on knowledge distillation in Section~\ref{sec:comparison results} followed by results on transfer learning in Section~\ref{sec:tl}. Finally, in Section~\ref{sec:analysis}, we analyse and discuss the impact of various hyper-parameters and design choices of our approach.

\subsection{Comparison with prior work} \label{sec:comparison results}

We extensively evaluate our approach on three different datasets: ImageNet~\cite{deng2009imagenet}, CIFAR-100~\cite{krizhevsky2009learning_cifar} and CIFAR-10.
We start with ImageNet, as this is the most challenging and interesting dataset among the three.
Then, on CIFAR-100 we conduct an
extensive evaluation on many network architectures and compare against several state-of-the-art methods.
Finally, we conclude with CIFAR-10. 

For all the experiments we use $\alpha=1$, $\beta=1$, and for ImageNet and CIFAR-100 $K=4096$, $\tau=0.2$ while for CIFAR-10 $K=256$, $\tau=0.005$. 
We apply our distillation loss on the feature maps of the last convolutional layer.
For complete implementation details, see Sections \ref{sec:appendix_impl} and \ref{sec:appendix_training} of Appendix.

\subsubsection{ImageNet results} \label{sec:imagenet_results}
\begin{table*}[t!]
    \centering
    \caption{\textbf{Evaluation on ImageNet.} Top-1 and Top-5 error rate of student network on ImageNet validation set, for two teacher-student combinations, ResNet34 (21.8M) to ResNet18 (11.7M) and ResNet50 (25.6M) to MobileNet (4.2M).  
    The results for other methods are taken from \cite{tian2019contrastive} and \cite{heo2019comprehensive}. FT refers to~\cite{kim2018paraphrasing}
    }
    \label{tab:imagenet}
    \ra{1.2}
    {\setlength{\extrarowheight}{1pt}\scriptsize{
    \begin{tabular}{c!{\hspace{0.4em}}c!{\hspace{0.4em}}c!{\hspace{0.4em}}|!{\hspace{0.4em}}c!{\hspace{0.4em}}c!{\hspace{0.4em}}c!{\hspace{0.4em}}c!{\hspace{0.4em}}c!{\hspace{0.4em}}c!{\hspace{0.4em}}c}\toprule
    & ResNet34 & ResNet18 & KD & AT+KD & SP & CC & 
    CRD & CRD+KD & Ours\\ 
    \cmidrule{2-10}
    Top-1 & 26.69 & 30.25 & 29.34 & 29.30 & 29.38 &
    30.04 & 28.83 & 28.62 &\textbf{28.33}\\
    Top-5 & 8.58 & 10.93 & 10.12 & 10.00 & 10.20 & 
    10.83 & 9.87 & 9.51 & \textbf{9.33}\\
    \bottomrule
    \\
    \cmidrule[\heavyrulewidth]{1-9}
    & ResNet50 & MobileNet & KD & AT+KD & FT & AB+KD & OFD & Ours &  \\ 
    \cmidrule{2-9}
    Top-1 & 23.87 & 31.13 & 31.42 & 30.44 & 30.12 & 31.11 & 28.75 & \textbf{27.46}  & \\
    Top-5 & 7.14 & 11.24 & 11.02 & 10.67 & 10.50 & 11.29 & 9.66 & \textbf{8.87} & \\
    \cmidrule[\heavyrulewidth]{1-9}
    \end{tabular}}}
\end{table*}

\input{cifar100_same_extended}
\input{cifar100_diff_extended}

In Table~\ref{tab:imagenet} we provide results for the ImageNet dataset, which contains 1.28M training images and 50K test images, over 1000 semantic classes.
It is much more challenging than CIFAR-100 and CIRAR-10,
which makes it the ultimate benchmark for evaluating distillation methods.

We evaluate our approach on it with two teacher-student combinations:
ResNet34 to ResNet18 (same architecture design) and ResNet50 to MobileNet \cite{howard2017mobilenets} (different architecture design). 
We observe that our distillation method reduces the Top-1 error of ResNet18 from $30.25\%$ to $28.33\%$ and MobileNet from $31.13\%$ to $27.46\%$, which sets the new state-of-the-art on this challenging dataset.

For ResNet34 to ResNet18, we use the results reported in CRD \cite{tian2019contrastive} for all other methods. Our approach outperforms all including CRD+KD which is second to ours with $0.30$ higher error rate. 
For ResNet50 to MobileNet, we follow OFD \cite{heo2019comprehensive}. We outperform OFD with $1.29\%$ lower error rate, which is a very significant improvement. Note that OFD extensively studied distillation with feature regression and carefully picks the location for feature regression, uses a modified ReLU (margin-ReLU) activation function, and 
a more robust (partial $L_2$) distance loss.
Nevertheless, our simple approach outperforms this well-engineered direct feature regression method.

Note that, due to the time-consuming nature of the ImageNet experiments, we did not try to tune the hyper-parameters of our method in this case (instead we reused the ones chosen for CIFAR-100). As a result, a further reduction of the error rates of the student might very well be possible by a more proper adjustment of the hyper-parameters.

\subsubsection{CIFAR-100 results} \label{sec:cifar_100_results}

The CIFAR-100 dataset is one of the most commonly used dataset for evaluating knowledge distillation methods.
It consists of small $32\times32$ resolution images and 100 semantic classes.
It has 50K images in the training set and 10K in the test set which are evenly distributed across the semantic classes. 
 
In Tables~\ref{tab:cifar100_same} and \ref{tab:cifar100_diff} we provide an exhaustive evaluation of our method under many different network architectures.
Also, we compare against several prior methods, i.e., KD~\cite{hinton2015distilling}, FitNet~\cite{romero2014fitnets}, AT \cite{zagoruyko2017AT}, AB~\cite{heo2019knowledge}, FSP~\cite{yim2017gift}, SP~\cite{tung2019SP_iccv}, VID~\cite{ahn2019variational}, and CRD~\cite{tian2019contrastive}. For network sizes and compression rates see \S \ref{sec:cifar100_compress} of Appendix. 

\paragraph{Knowledge transfer between networks with the same architecture design.} 
In Table \ref{tab:cifar100_same}, we compare knowledge distillation methods with teacher and student networks that have the same architecture design but different depth or width (width refers to the number of channels per layer).
For the network architectures, we evaluate knowledge distillation between WideResNet \cite{zagoruyko2016WRN}, ResNet \cite{he2016deep} and VGG \cite{simonyan2014very}.

Our method outperforms all the other methods on all the different teacher-student combinations, except for the \textit{VGG13 to VGG8} case where we are second only to CRD \cite{tian2019contrastive}~
 a contemporary method to ours. 
On average we improve by an absolute $2.97\%$ over students without distillation,\textit{i.e.}
relatively $16.9\%$ more than CRD ($2.54\%$ over student) and $7.6\%$ more than CRD+KD ($2.76\%$).
Moreover, in some cases our students achieve accuracy 
either very close to that of the teacher (\textit{ResNet56 to ResNet20} and \textit{ResNet110 to ResNet32}), or even exceeds it (\textit{WRN-40-2 to WRN-40-1}), which further confirms the potency of our method.

\paragraph{Knowledge transfer between networks with different architectures.} 
In Table \ref{tab:cifar100_diff}, we evaluate the merit of our approach for 
distillation between different network designs. All of them but one have different spatial dimensions between feature maps of student and teacher. 
For example, MobileNetV2 at the penultimate layer has $2\times2$ feature maps while the teachers, VGG13 and ResNet50, have $4\times4$. Similarly, ShuffleNetV1/V2 have $4\times4$ while ResNet32x4 and WRN-40-2 have $8\times8$. As the 
soft-assignment maps $\T{\mm p}$ and $\s{\mm p}$ should have the same spatial dimension to apply our distillation loss, we do average pooling on the feature maps of teacher before quantization, as explained in section~\S\ref{sec:predict_quantizations}. 
 
We observe that our approach outperforms every other method on all but one experiment. 
In the cases of the ResNet32x4 teacher, we notice an improvement of more than $1\%$ against the other methods. 
Overall, the proposed approach improves by an average of $5.25\%$ on the student without distillation. 
While the most competitive CRD and CRD+KD bring respectively an improvement of $4.59\%$ and $4.85\%$. 
Note that, in terms of average gain over the student without distillation, we get a relative improvement of $14.29\%$ and $8.21\%$ compared to CRD and CRD+KD respectively.

\input{cifar10.tex}

\subsubsection{CIFAR-10 results} \label{sec:cifar_10_results}
The CIFAR-10 dataset is similar to CIFAR-100 with the  only difference that there are now $10$ semantic classes instead of $100$.

\paragraph{Comparison with prior work.} 
In Table \ref{tab:cifar10}, we compare our approach with KD, AT and SP in terms of error rate on CIFAR-10 test set. 
We consider distillation between WideResNet student and teacher with different depth and/or width. 
Again, our method achieves state-of-the-art results on CIFAR-10. 
Specifically, we outperform the other methods on two settings (\textit{WRN-40-1} to \textit{WRN-16-1} and \textit{WRN-16-8} to \textit{WRN-16-2}), while we achieve almost the same results on the other three settings with statistically negligible difference of less than $0.04\%$.
Our approach achieves an average reduction in error rate of $0.82\%$ compared to the student without distillation, 
while the most competitive SP method gets $0.75\%$ followed by AT with $0.48\%$.

\subsection{Transfer learning to small-sized datasets}\label{sec:tl}
\input{transfer_learning}

In this section, we evaluate our method for transfer learning. In transfer learning, the objective is to train a student on a small target dataset with the aid of a teacher which is 
pre-trained on a large dataset. For our experimental evaluation, we use ResNet34 pre-trained on ImageNet as the teacher and, ResNet18 and VGG-9 \cite{simonyan2014very} as the student network. We use MIT-67 \cite{quattoni2009recognizing_mit67} as the target dataset. It contains $15,620$ indoor scene images, classified into $67$ classes. Following the evaluation protocol of \cite{ahn2019variational}, we sub-sample the training set with $M=\{80, 50, 25, 10\}$ images per classes. This is to assess the performance at various levels of availability of the training data. The student is trained from scratch on target data with cross-entropy loss and distillation loss.
For all the transfer learning experiments we use $K=4096$ visual words which is 
learned on ImageNet, where the teacher network 
was pre-trained.

In case of transfer learning we found that it is better to apply the proposed loss at the last two layers (layer3 and layer4) of ResNet34. 
For student network ResNet18 we use the same, layer3 and layer4, and for VGG-9 we use the last two max-pool layers.
Exactly the same two feature levels are being used by the competing methods in this section.

The results for transfer learning experiments are given in Table \ref{tab:tl}. The table compares our method to several distillation approaches including FitNet, AT, NST VID. In the table, LwF refers to learning without forgetting \cite{li2017learning}, VID-LP and VID-I are VID \cite{ahn2019variational} loss on logits and on intermediate features respectively, while VID-LP+VID-I uses both. 
Our proposed method outperforms all the methods on 3 out of 4 configurations of $M$ for both the students. While being second only to VID-I with a statistically insignificant difference of less than $0.1\%$.
 
We also give results with the fine-tuning case, 
i.e., pre-training on ImageNet and then fine-tuning on MIT-67 without any distillation loss. 
Fine-tuning is a very strong baseline and standard method for transfer learning. 
However, it is not very practical since it requires pre-training the student on a large-size dataset (which is computationally expensive and needs access and storage of the dataset).
In our experiments, the proposed method performs better than even fine-tuning.

\subsection{Further analysis} \label{sec:analysis}

\begin{figure}[t!]
    \begin{minipage}{.5\textwidth}
    \centering
    \includegraphics[width=0.9\textwidth]{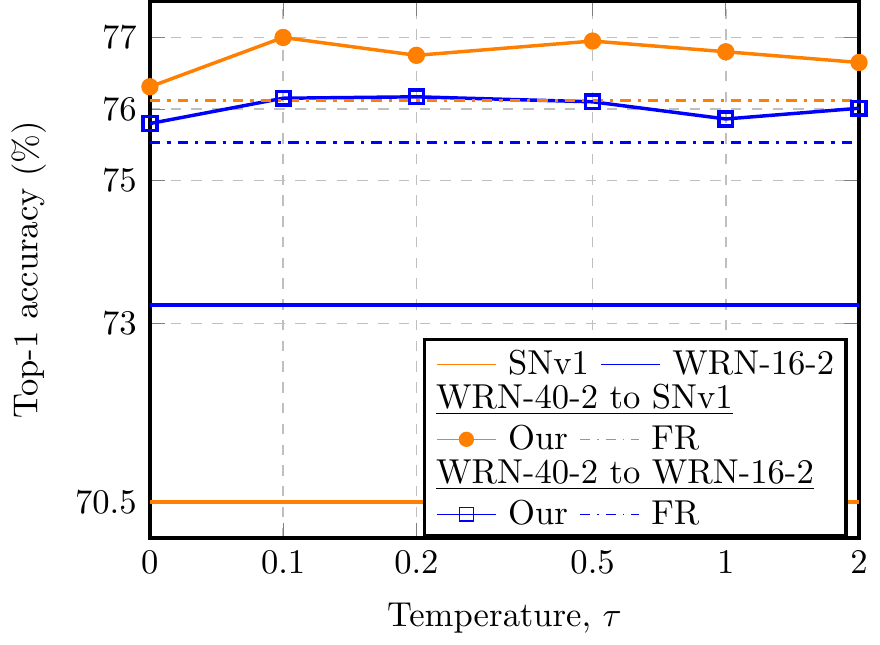}
    \subcaption{Accuracy vs.\ $\tau$}\label{fig:tau}
    \end{minipage}
    \begin{minipage}{.5\textwidth}
    \centering
    \includegraphics[width=0.9\textwidth]{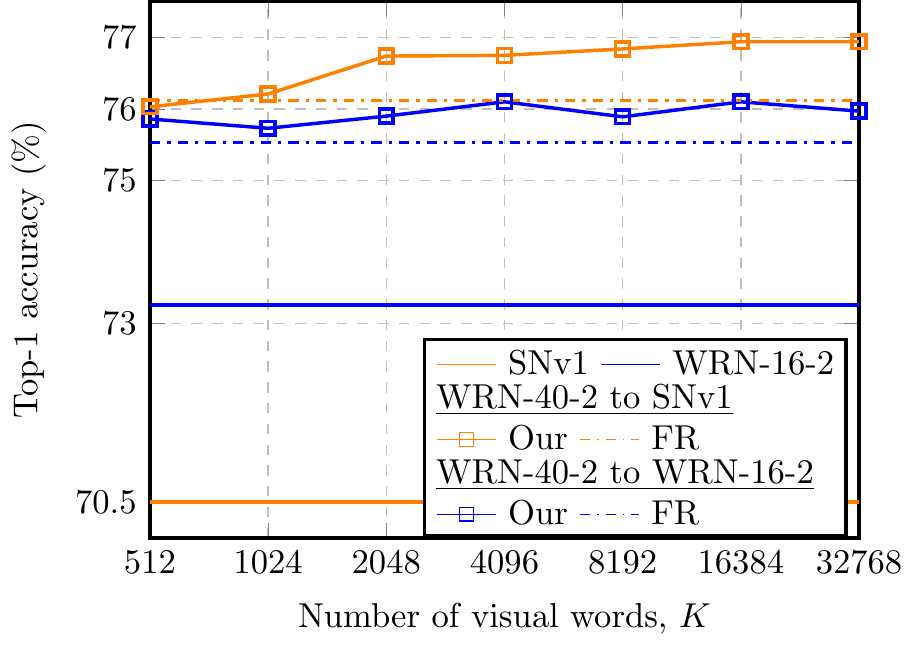}
    \subcaption{Accuracy vs.\ $K$}\label{fig:vary_k}
    \end{minipage}
    \caption{\textbf{Effect of varying temperature $\tau$ and visual words $K$}. 
    Top-1 accuracy on CIFAR-100 of students trained with proposed distillation loss with varying hyperparameters $\tau$ and $K$.
    The graphs show performance of two students, WRN-16-2 and ShuffleNetv1 (SNv1) with WRN-40-2 as teacher. 
    (a) Effect of temperature $\tau$. (b) Effect of varying number of visual words.
    The straight lines refer to the student trained without distillation (solid line) or with feature regression (FR, dash-dotted line)
    }
    \label{fig:vary_k_t}
\end{figure}

Here we analyse several aspects of our method for the model compression scenario.

\paragraph{Impact of temperature value $\tau$.}
In Figure~\ref{fig:tau} we plot how the temperature value $\tau$ of the soft-quantization of the teacher feature maps in Eq.\ (\ref{eq:softquant}) affects the performance of our  method.
We provide results for two teacher-student configurations, WRN-40-2 to WRN-16-2 and WRN-40-2 to ShuffleNetV1, and 5 different $\tau$ values, $0.1$, $0.2$, $0.5$, $1.0$, and $2.0$.
We also provide results for the hard-assignment case, which we denote with the $\tau=0$ in the plot.
We observe that choosing a small $\tau$ value, which means a more peaky soft-assignment (i.e., closer to hard-assignment), leads to better distillation performance. 
However, going to the extreme case of hard-assignment (i.e., $\tau=0$) drops the distillation performance.
Hence, very peaky soft-assignment achieves better knowledge transfer than the hard-assignment case.

\paragraph{Impact of vocabulary size $K$.}
In Figure~\ref{fig:vary_k} we plot the distillation performance of our method as a function of the vocabulary size $K$ on CIFAR-100.
We observe a relatively stable performance for $K\ge 2048$. In our CIFAR-10 experiments we noticed that $K$ between $128$ to $256$ was leading to better performance\footnote{See Fig.\,\ref{fig:c10_k} in Sec.\,\ref{sec:app_cifar10} of Appendix for performance-vs.-$K$ plot on CIFAR-10.}. 
Therefore, it seems that the number of visual words $K$ depends on the number of classes, preferably it should be sufficiently larger than the number of classes.

\begin{table}[t!]
    \centering
    \ra{1.2}
        \caption{\textbf{Distillation at different layers.} Comparing student networks with the proposed approach when applied at different layers. Block $L$ corresponds to the penultimate layer (before classification layer) while Block $L-1$ refers to the layer at a block before Block $L$. We report the Top-1 accuracy
        }
    \label{tab:diff_layer}
    {\setlength{\extrarowheight}{1pt}\scriptsize{
    \begin{tabular}{l!{\hspace{0.4em}}c!{\hspace{0.4em}}c!{\hspace{0.4em}}l!{\hspace{0.4em}}c} \toprule
     \multicolumn{2}{c}{CIFAR-100}&& \multicolumn{2}{c}{CIFAR-10} 
     \\ 
     \cmidrule{1-2} \cmidrule{4-5}
     \multicolumn{2}{c}{WRN-40-2$\rightarrow$WRN-16-2}&& \multicolumn{2}{c}{WRN-40-1$\rightarrow$WRN-16-1}
     \\ \midrule
     Block $L-1$ & 74.16 && Block $L-1$ & 91.64
     \\
     Block $L$ & \textbf{76.10} && Block $L$ & \textbf{91.98} 
     \\ \bottomrule
    \end{tabular}}}
\end{table}

\paragraph{At which feature level to apply the distillation loss?}
In Table~\ref{tab:diff_layer} we measure the performance of our method when the distillation loss is applied (1) to the last feature level of the teacher-student networks 
(which is what we have used for all other experiments on the model compression scenario),
and (2) to the feature level of the previous down-sampling stage (\ie, the output of the 2nd residual block in WRN-40-2/WRN-16-2 or WRN-40-1/WRN-16-1).
We report results on CIFAR-100 and CIFAR-10.
In all cases, switching to the feature maps of the previous down-sampling stage leads to a drop in performance.
Therefore, our proposed distillation loss appears to perform best with the last feature level.

\input{fr_table}
\paragraph{Comparison with direct feature regression.} 
In Table \ref{tab:feat_reg}, 
we explicitly compare our approach against the baseline method of directly regressing raw feature maps (Feature regression).
To that end, we apply the Feature regression distillation loss on the same feature layer as in our method, i.e., the last convolutional layer. 
Also, we train the student network under the same conditions as in our method.
In the upper part of the table, 
we evaluate distillation performance with different capacity networks. 
In this case, we add a $1\times1$ convolutional regression layer on top of the chosen layer for distillation of the student network.
This is needed to match feature dimension of the teacher. 
Not surprisingly, our approach outperforms Feature regression for all the teacher-student pairs.
Further, we also evaluate on distillation between same architecture and same capacity networks. 
This is an ideal case for Feature regression as there is no burden on student to mimic a higher capacity network and also no need for additional regression layer. 
Even in this ideal setup for Feature regression we outperform it for three architectures while being marginally lower on the other two;
overall, our method achieves an average relative gain of $0.76\%$ over Feature regression.

%% file: cifar100_same_extended.tex
\begin{table*}[t!]
    \centering
    \caption{\textbf{Experiments on CIFAR-100.} Top-1 test accuracy of student networks for various student-teacher combinations and knowledge distillation methods. 
    The pre-trained teacher models are taken from \cite{tian2019contrastive} and for all methods except ours we use the results reported in \cite{tian2019contrastive} for a fair comparison. We report average over 5 runs as in \cite{tian2019contrastive}}
    \label{tab:cifar100_same}
    \ra{1.2}
    \resizebox{\linewidth}{!}{
    {\setlength{\extrarowheight}{1pt}\scriptsize{
    \begin{tabular}{ccc <{\hspace{0.4em}} c <{\hspace{0.4em}} c <{\hspace{0.4em}} c <{\hspace{0.4em}} c <{\hspace{0.4em}} c <{\hspace{0.4em}} c <{\hspace{0.4em}} c <{\hspace{0.4em}} c <{\hspace{-0.1em}} c <{\hspace{-0.1em}} c <{\hspace{0.4em}} c <{\hspace{0.4em}} c}\toprule
    \multicolumn{2}{c}{Model} && \multicolumn{10}{c}{Knowledge distillation methods} &\\ 
    \cmidrule{4-13}
         Teacher & Student && KD & FitNet & AT & AB & FSP & SP & VID & CRD & CRD+KD & Ours & teacher & student \\ \midrule
         WRN-40-2 & WRN-16-2 &
         & 74.92 & 73.58 & 74.08 & 72.50 & 72.91 & 73.83 & 74.11 &  75.48 & 75.64 &  \textbf{76.10} & 75.61 & 73.26\\
         WRN-40-2 & WRN-40-1 &
         & 73.54 & 72.24 & 72.77 & 72.38 & - & 72.43 & 73.30 & 74.14 & 74.38 &
         \textbf{74.58} & 75.61 & 71.98 \\ 
         ResNet56 & ResNet20 &
         & 70.66	&	69.21	&	70.55	& 69.47 & 69.95 & 69.67	&	70.38	&	71.16	&	71.63 & \textbf{71.84} & 72.34 & 69.06 \\ 
         ResNet110 & ResNet20&
         & 70.67	&	68.99	&	70.22	& 69.53	& 70.11 & 70.04	&	70.16	&	71.46	&	71.56 & \textbf{71.89} & 74.31 & 69.06 \\ 
         ResNet110 & ResNet32&
         & 73.08	&	71.06	&	72.31	& 70.98 & 71.89 & 72.69	&	72.61	&	73.48	&	73.75	& \textbf{74.08} & 74.31 & 71.14 \\ 
         ResNet32x4 & ResNet8x4&
         & 73.33	&	73.50	&	73.44	& 73.17 & 72.62 & 72.94	&	73.09	&	75.51	&	75.46	& \textbf{75.88} & 79.42 & 72.50 \\ 
         VGG13 & VGG8&
         & 72.98	&	71.02	&	71.43	& 70.94 & 70.23 & 72.68	&	71.23	&	73.94	&	\textbf{74.29}	& 73.81 & 74.64 & 70.36 \\ \bottomrule
    \end{tabular}}}}
    
\end{table*}

%% file: cifar100_diff_extended.tex
\begin{table*}[ht]
    \centering
    \caption{\textbf{Distillation between different architectures.} Top-1 test accuracy on CIFAR100 dataset of the student networks. The student models are learned with knowledge distillation from a teacher with different architecture. The table compares various distillation methods. Similar to Table \ref{tab:cifar100_same}, we use the pre-trained teacher networks provided by \cite{tian2019contrastive} and the results for other methods are also taken from \cite{tian2019contrastive}. Following the protocol of \cite{tian2019contrastive}, we report average over 3 runs}
    \label{tab:cifar100_diff}
       \ra{1.2}
    \resizebox{\linewidth}{!}{
    {\setlength{\extrarowheight}{1pt}\scriptsize{
    \begin{tabular}{c <{\hspace{0.4em}} >{\hspace{0.4em}} c >{\hspace{0.4em}} c >{\hspace{0.4em}} c >{\hspace{0.4em}} c >{\hspace{0.4em}} c >{\hspace{0.4em}} c >{\hspace{0.4em}} c >{\hspace{0.4em}} c >{\hspace{0.4em}} c >{\hspace{0.4em}} c >{\hspace{0.4em}} c >{\hspace{0.4em}} c >{\hspace{0.4em}} c}\toprule
    \multicolumn{2}{c}{Model} && \multicolumn{9}{c}{Knowledge distillation methods
} &\\
    \cmidrule{4-12}
         Teacher & Student && KD & FitNet & AT & AB & SP & VID & CRD & CRD+KD & Ours & teacher & student \\ \midrule
         VGG13 & MobileNetV2 &
         & 67.37	&	64.14	&	59.4	& 66.06  & 66.3	&	65.56	&	69.73	&	\textbf{69.94}	& 68.79 & 74.64 & 64.6 \\ 
         ResNet50 & MobileNetV2 &
         & 67.35	&	63.16	&	58.58	& 67.20 & 68.08	&	67.57	&	69.11	&	69.54	& \textbf{69.81} & 79.34 & 64.6 \\ 
         ResNet50 & VGG8 &
         & 73.81	&	70.69	&	71.84	& 70.65 & 73.34	&	70.30	&	74.30	&	74.58	& \textbf{75.17} & 79.34 & 70.36 \\
         ResNet32x4 & ShuffleNetV1 &
         & 74.07	&	73.59	&	71.73	& 73.55 & 73.48	&	73.38	&	75.11	&	75.12	& \textbf{76.28} & 79.42 & 70.50 \\
         ResNet32x4 & ShuffleNetV2 &
         & 74.45	&	73.54	&	72.73	& 74.31 & 74.56	&	73.4	&	75.65	&	76.05	& \textbf{77.09} & 79.42 & 71.82  \\
         WRN-40-2 & ShuffleNetV1 &
         & 74.83	&	73.73	&	73.32	& 73.34 & 74.52	&	73.61	&	76.05	&	76.27	& \textbf{76.75} & 75.61 & 70.50 \\ \bottomrule
    \end{tabular}}}}
    
\end{table*}

%% file: cifar10.tex
\begin{table*}[t!]
    \centering
    \caption{\textbf{CIFAR-10 experiments.} Top-1 error of student networks on CIFAR-10. We use the results reported in \cite{tung2019SP_iccv} and following it, use median of 5 runs}
    \label{tab:cifar10}
    \ra{1.2}
    {\setlength{\extrarowheight}{1pt}\scriptsize{
    \begin{tabular}{c!{\hspace{0.4em}}c!{\hspace{0.4em}}c!{\hspace{0.4em}}c!{\hspace{0.4em}}c!{\hspace{0.4em}}c!{\hspace{0.4em}}c!{\hspace{0.4em}}c!{\hspace{0.4em}}c!{\hspace{0.4em}}c}\toprule
    \multicolumn{2}{c}{Model} && \multicolumn{4}{c}{Knowledge Distillation methods} &&\\
    \cmidrule{4-7}
    Teacher ($\#$params) & Student ($\#$params) && KD & AT+KD & SP & Ours && teacher & student\\ \midrule
    WRN-40-1 (0.56M) & WRN-16-1 (0.17M)  && 8.48 & 8.30 & 8.13 & \textbf{8.02} && 6.51 & 8.74 \\
    WRN-16-2 (0.69M) & WRN-16-1 (0.17M) && 7.94 & 8.28 & \textbf{7.52} & 7.55 && 6.07 & 8.74 \\ 
    WRN-40-2 (2.24M) & WRN-16-2 (0.69M) && 6.00 & 5.89 & \textbf{5.52} & 5.56 && 5.18 & 6.07 \\
    WRN-16-8 (11M) & WRN-16-2 (0.69M) && 5.62 & 5.47 & 5.34 & \textbf{5.06} && 4.24 & 6.07 \\
    WRN-16-8 (11M) & WRN-40-2 (2.24M) && 4.86 & \textbf{4.47} & 4.55 & 4.48 && 4.24 & 5.18\\ \bottomrule
    \end{tabular}}}

\end{table*}

%% file: transfer_learning.tex
\begin{table}[t!]
    \centering
    \caption{\textbf{Results for transfer learning}. The teacher is pre-trained on ImageNet while the student is trained from scratch on a target dataset with cross-entropy loss and distillation loss. 
    The results for other methods are reported from \cite{ahn2019variational}}
    \label{tab:tl}

    \begin{subtable}{.49\linewidth}\centering{
    \caption{MIT-67, ResNet34 to ResNet18}
    {\setlength{\extrarowheight}{1pt}\scriptsize{
    \begin{tabular}{ccccccccc}
    \toprule
    \multicolumn{1}{r}{M } & 80 & & 50 & & 25 & & 10\\
    \midrule
    \multicolumn{1}{l}{student} & 48.13 & & 37.69 & & 27.01 & & 14.25\\
    \multicolumn{1}{l}{fine-tuning} & 70.97 & & 66.04 & & 58.13 & & 47.91 \\
    \midrule
    \multicolumn{1}{l}{LwF} & 63.43 & & 51.79 & & 41.04 & & 22.76 \\
    \multicolumn{1}{l}{FitNet}& 71.34 & & 60.45 & & 54.78 & & 36.94 \\
    \multicolumn{1}{l}{AT} & 58.21 & & 48.66 & & 43.66 & & 27.01\\
    \multicolumn{1}{l}{NST} & 55.52 & & 46.34 & & 33.21 & & 20.82\\
    \multicolumn{1}{l}{VID-LP} & 67.91 & & 58.51 & & 47.09 & & 31.94\\
    \multicolumn{1}{l}{VID-I} & 71.34 & & 63.66 & & 60.07 & & \textbf{50.97}\\
    \multicolumn{1}{l}{LwF+FitNet} & 70.97 & & 60.37 & & 54.48 & & 38.73\\
    \multicolumn{1}{l}{VID-LP+VID-I} & 71.87 & & 65.75 & & 61.79 & & 50.37\\
    \midrule
    \multicolumn{1}{l}{Ours} & \textbf{73.18} & & \textbf{69.40} & & \textbf{62.71} & & 50.92 \\
    \bottomrule
    \end{tabular}}}}
    \end{subtable}
    \hfill
    \begin{subtable}{.49\linewidth}\centering{
    \caption{MIT-67, ResNet34 to VGG-9}
    {\setlength{\extrarowheight}{1pt}\scriptsize{
    \begin{tabular}{cccccccc}
    \toprule
    \multicolumn{1}{r}{M } & 80 & & 50 & & 25 & & 10\\
    \midrule
    \multicolumn{1}{l}{student} & 53.58 & & 43.96 & & 29.70 & & 15.97 \\
    \multicolumn{1}{l}{fine-tuning} & 65.97 & & 58.51 & & 51.72 & & 39.63 \\
    \midrule 
    \multicolumn{1}{l}{LwF} & 60.90 & & 52.01 & & 41.57 & & 27.76 \\
    \multicolumn{1}{l}{FitNet} & 70.90 & & 64.70 & & 54.48 & & 40.82 \\
    \multicolumn{1}{l}{AT} & 60.90 & & 52.16 & & 42.76 & & 25.60 \\
    \multicolumn{1}{l}{NST} & 55.60 & & 46.04 & & 35.22 & & 21.64\\
    \multicolumn{1}{l}{VID-LP} & 68.88 & & 61.64 & & 50.22 & & 39.25 \\
    \multicolumn{1}{l}{VID-I} & \textbf{72.01} & & 67.01 & & 59.33 & & 45.90 \\
    \multicolumn{1}{l}{LwF+FitNet} & 70.52 & & 64.10 & & 54.63 & & 40.15 \\
    \multicolumn{1}{l}{VID-LP+VID-I} & 71.72 & & 66.49 & & 58.96 & & 45.89 \\
    \midrule
    \multicolumn{1}{l}{Ours} & 71.92 & & \textbf{67.79} & & \textbf{60.10} & & \textbf{47.99} \\
    \bottomrule
    \end{tabular}}}}
    \end{subtable}

\end{table}

%% file: fr_table.tex
\begin{table}
    \centering
    \caption{
    \textbf{Comparison with direct feature regression on CIFAR-100.} 
    Upper part of the table shows distillation performance between different capacity networks, thus we use a regression layer on student for the Feature regression method.
    In the lower part, distillation is between same architecture and same capacity networks thus we do not use any additional
    layer to regress.
    Relative gain refers to the relative improvement of our method over Feature regression
    }
    \label{tab:feat_reg}
    \setlength{\extrarowheight}{1pt}\scriptsize{
    \begin{tabular}{c!{\hspace{0.4em}}c!{\hspace{0.4em}}c!{\hspace{0.4em}}c!{\hspace{0.4em}}c!{\hspace{0.4em}}c!{\hspace{0.4em}}c!{\hspace{0.4em}}c!{\hspace{0.4em}}c}
    \toprule
    Teacher & Student & & student & & Feature regression && Ours & relative gain (\%) \\
    \midrule
    ResNet50 & VGG8 && 70.36 && 71.9 && \textbf{75.17} & 4.55\\
    WRN-40-2 & WRN-16-2 &&  73.26 && 75.53 && \textbf{76.10} & 0.75\\
    WRN-40-2 & ShuffleNetV1 && 70.50 && 75.89 && \textbf{76.75} & 1.13 \\
    ResNet32x4 & ResNet8x4 && 72.50 && 74.12 && \textbf{75.88} & 2.37\\
    ResNet32x4 & ShuffleNetV2 && 71.82 && 75.58 && \textbf{77.09} & 2.00\\
    ResNet56 & ResNet20 &&  69.06 && 71.56 && \textbf{71.84} & 0.39 \\
    \midrule
    ResNet50 & ResNet50 && 79.34 &&  79.12 && \textbf{80.58} & 1.85\\
    WRN40-2 & WRN40-2 && 75.61 && \textbf{78.26}  && 77.96 & -0.38\\
    ResNet110 & ResNet110 && 74.31 && 75.52  && \textbf{76.15} & 0.83\\
    ResNet32x4 & ResNet32x4 && 79.42 && \textbf{80.57}  && 80.53 & -0.05\\
    ResNet56 & ResNet56 && 72.34 && 73.64  && \textbf{74.77} & 1.53\\
    \bottomrule
    \end{tabular}}
    \vspace{-5pt}
\end{table}

%% file: appendix.tex
\section{On using cosine similarity in assignment predictor}\label{sec:appendix_cs}
In the assignment predictor we use cosine similarity to predict the $K$-dimensional soft-assignment vector $\s{\mv p^{(h, w)}}$ from the feature vector $\s{\mm{f}^{(h, w)}}$ (Eq. \ref{eq:apredictor}). 
The reason for choosing this similarity measure over the Euclidean distance is that the former 
L2-normalizes the features and the visual word weights, which we observed to lead to better behavior. We believe that this is due to the fact that the L2-normalization acts as a regularizer for the weights of the visual words in the assignment predictor in case of unbalanced k-means clusters: without the L2-normalization, more frequent teacher-words would lead to bigger weight magnitudes for the corresponding student-words. Also, with cosine similarity, it is easier to control the peakiness of the predicted word distribution since the range of its output values is fixed and a priori known (i.e., between -1 and 1).

\section{Implementation details for vector quantization}\label{sec:appendix_impl}

In our quantization-based distillation method we use k-means to learn the visual teacher-words vocabulary $V$.
Here we provide implementation details regarding how we apply the k-means clustering algorithm.

\textbf{k-means implementation.} For k-means, we use the implementation provided by the publicly available FAISS~\cite{JDH17} library.

\textbf{Applying k-means on ImageNet.}
The training set of ImageNet is quite large (i.e., it has around 1.28M images).
Therefore, when evaluating our distillation method on it, 
to learn efficiently the visual teacher-words vocabulary $V$, we apply k-means only to a randomly sampled subset of 0.2M images from this set. Given the spatial size of feature maps, this subset provides a sufficiently large corpus of vectors to learn a vocabulary of size $K=4096$, as we use in our experiments.

\textbf{Applying k-means on CIFAR-100 and CIFAR-10.}
For CIFAR-100 and CIFAR-10 experiments, we apply k-means to the entire training sets. 

\section{Training details}\label{sec:appendix_training}

\subsection{Model compression}

For the ImageNet experiments in Section \ref{sec:comparison results}, following \cite{tian2019contrastive} and \cite{heo2019comprehensive}, we train for $100$ epochs with the initial learning rate of $0.1$ which is reduced every $30$ epochs with a decay rate of $0.1$. For ResNet34 to ResNet18, we use a batch size of $256$, while for ResNet50 to MobileNet we use $210$ as batch size due to GPU memory constraints. For the hyper-parameters of our distillation method, we use $\alpha=1$, $\beta=1$, $\tau=0.2$ and $K=4096$.

For all the experiments on CIFAR-100, we follow the protocol of \cite{tian2019contrastive} for training the student networks. Specifically, in all cases we train the student for $240$ epochs with batch size of $64$ and an initial learning rate of $0.05$, which we drop by a factor of $0.1$ after $150$, $180$, and $210$ epochs. The only exception is MobileNetV2 and ShuffleNetV1/V2, as in \cite{tian2019contrastive}, where the learning rate is initialized to $0.01$. The hyper-parameters of our method are $\alpha=1$, $\beta=1$, $\tau=0.2$ and $K=4096$.

For the CIFAR-10 experiments we follow the protocol of \cite{tung2019SP_iccv} and train the student for $200$ epochs with a batch size of $128$.
The initial learning rate is set to $0.1$ which decays by a factor of $0.2$ at $60^{\text{th}}$, $120^{\text{th}}$, $160^{\text{th}}$ epoch. The hyper-parameters of our distillation method are set to $\alpha=1$, $\beta=1$, $\tau=0.005$ and $K=256$.

\subsection{Transfer learning to small-sized datasets}

For the transfer learning experiments in Section\,\ref{sec:tl} of the main paper, we used the hyper-parameters $\alpha=1$, $K=4096$, and $\tau$ equal to $0.2$ and $0.002$ for layer4 and layer3 of ResNet34 respectively (for both layers the $\tau$ value was chosen so that, as mentioned in the main paper, the softmax probability for the closest visual teacher-word is on average around 0.996).
We found that in the transfer learning experiments it is important to tune properly the $\beta$ hyper-parameter so as to prevent overfitting on the classification task of the training images. 
To that end, as it is recommended in the evaluation protocol of~\cite{ahn2019variational}, we used $20\%$ of the training images as validation images and we tuned the $\beta$ hyper-parameter on them. Specifically, for the ResNet18 student experiments we used $\beta=10.0$ and for the VGG9 student experiments we used $\beta=20.0$.
To train the students we follow the training protocol of~\cite{ahn2019variational},
\ie, $200$ training epochs with an initial learning rate of $0.05$ which is dropped by a factor of $10$ after $150$ epochs.
The batch size is $128$ for the ResNet18 student and $32$ for the VGG9 student. 

\section{Model compression in CIFAR-100 experiments}\label{sec:cifar100_compress}
Table \ref{tab:cifar100_nets} gives the number of parameters of all the networks used in Tables \ref{tab:cifar100_same} and \ref{tab:cifar100_diff} for CIFAR-100 experiments. The table shows that we can get high reduction in parameters with significantly less drop in performance with our proposed method of knowledge distillation. In case of WRN-40-2 to WRN-16-2 and to ShuffleNetV1, we even get an improvement of $0.49\%$ and $1.14\%$ in accuracy over the teacher with compression of $68.81\%$ and $57.91\%$ respectively.

\begin{table}
    \centering
        \caption{Number of parameters of the teacher and student networks used in CIFAR-100 experiments and compression obtained by replacing the teacher with the student network. The compression is computed as percentage of reduction in number of parameters with respect to the teacher network}
    \label{tab:cifar100_nets}
    \begin{tabular}{llcccc}
    \toprule
    \multicolumn{2}{c}{Model}& &\multicolumn{3}{c}{Accuracy}\\
    \cmidrule(r{8pt}){1-2} \cmidrule(lr){4-6}
    Teacher network & Student network & compression (\%) & teacher &  student & Ours \\
    \midrule
    WRN-40-2 (2.26M) & WRN-16-2 (0.70M) & 68.81 & 75.61 & 73.26 & 76.10 \\
    WRN-40-2 (2.26M) & WRN-40-1 (0.57M) & 74.73 & 75.61 & 71.98 & 74.58\\
    ResNet56 (0.86M) & ResNet20 (0.28M) & 67.70 & 72.34 & 69.06 & 71.84\\
    ResNet110 (1.73M) & ResNet20 (0.28M) & 83.97 & 74.31 & 69.06 & 71.89\\
    ResNet110 (1.73M) & ResNet32 (0.47M) & 72.78 & 74.31 & 71.14 & 74.08\\
    ResNet32x4 (7.43M) & ResNet8x4 (1.23M) & 83.41 & 79.42 & 72.50 & 75.88\\
    VGG13 (9.46M) & VGG8 (3.96M) & 58.10 & 74.64 & 70.36 & 73.81\\
    VGG13 (9.46M) & MobileNetV2 (0.81M) & 91.41 & 74.64 & 64.6 & 68.79\\
    ResNet50 (23.7M) & MobileNetV2 (0.81M) & 96.57 & 79.34 & 64.6 & 69.81\\
    ResNet50 (23.7M) & VGG8 (3.96M) & 83.27 & 79.34 & 70.36 & 75.17\\
    ResNet32x4 (7.43M) & ShuffleNetV1 (0.95M) & 87.23 & 79.42 & 70.50 & 76.28\\
    ResNet32x4 (7.43M) & ShuffleNetV2 (1.36M) & 81.77 & 79.42 & 71.82 & 77.09\\
    WRN-40-2 (2.26M) & ShuffleNetV1 (0.95M) & 57.91 & 75.61 & 70.50 & 76.75\\
    \bottomrule
    \end{tabular}
\end{table}

\begin{figure}
    \centering
    \includegraphics[width=0.49\textwidth]{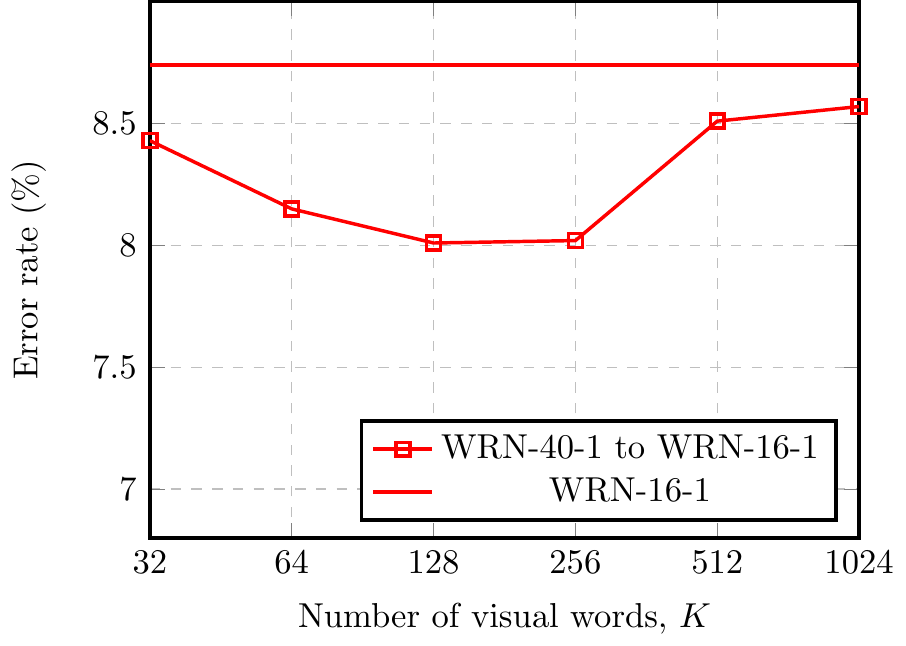}
    \caption{\textbf{Effect of varying $K$}. Error rate vs. $K$ on CIFAR-10 with WRN-16-1 as the student networks. The students are trained on the proposed distillation loss with WRN-40-1 as the teacher with varying number $K$ of visual teacher-words. The solid straight line represents student trained without distillation loss}
    \label{fig:c10_k}
\end{figure}

\section{Effect of vocabulary size on CIFAR-10 experiments}\label{sec:app_cifar10}

In Section \ref{sec:analysis} of the main paper, we discussed the impact of teacher vocabulary size $K$, based on the plot of accuracy versus $K$ for CIFAR-100 in Figure \ref{fig:vary_k}. Here we provide the analogous plot for CIFAR-10 in Figure \ref{fig:c10_k}. As we mentioned in the paper, $K$ between $128$ to $256$ leads to better performance.

\section{Qualitative results: Alignment of teacher and student quantized feature maps}

\begin{figure*}[t!]
\fboxsep=0pt
\fboxrule=1.5pt
    \centering
    \begin{tabular}{ccccccccccccc}\toprule
    Query && \multicolumn{10}{c}{Retrieved images}\\ \midrule
    \includegraphics[width=0.08\textwidth]{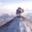} && \fcolorbox{red}{red}{\includegraphics[width=0.08\textwidth]{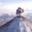}} & \includegraphics[width=0.08\textwidth]{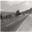} & \includegraphics[width=0.08\textwidth]{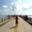} & \includegraphics[width=0.08\textwidth]{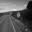}& \includegraphics[width=0.08\textwidth]{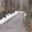} & \includegraphics[width=0.08\textwidth]{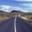}& \includegraphics[width=0.08\textwidth]{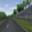} & \includegraphics[width=0.08\textwidth]{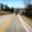}& \includegraphics[width=0.08\textwidth]{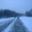} & \includegraphics[width=0.08\textwidth]{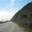} \\
    \includegraphics[width=0.08\textwidth]{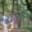} && \includegraphics[width=0.08\textwidth]{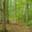} & \fcolorbox{red}{red}{\includegraphics[width=0.08\textwidth]{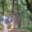}} & \includegraphics[width=0.08\textwidth]{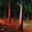} & \includegraphics[width=0.08\textwidth]{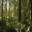}& \includegraphics[width=0.08\textwidth]{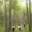} & \includegraphics[width=0.08\textwidth]{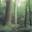}& \includegraphics[width=0.08\textwidth]{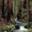} & \includegraphics[width=0.08\textwidth]{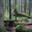}& \includegraphics[width=0.08\textwidth]{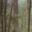} & \includegraphics[width=0.08\textwidth]{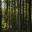} \\
    \includegraphics[width=0.08\textwidth]{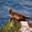} && \includegraphics[width=0.08\textwidth]{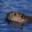} & \fcolorbox{red}{red}{\includegraphics[width=0.08\textwidth]{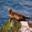}} & \includegraphics[width=0.08\textwidth]{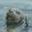} & \includegraphics[width=0.08\textwidth]{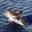}& \includegraphics[width=0.08\textwidth]{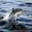} & \includegraphics[width=0.08\textwidth]{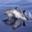}& \includegraphics[width=0.08\textwidth]{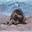} & \includegraphics[width=0.08\textwidth]{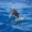}& \includegraphics[width=0.08\textwidth]{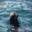} & \includegraphics[width=0.08\textwidth]{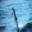} \\
    \includegraphics[width=0.08\textwidth]{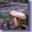} && \includegraphics[width=0.08\textwidth]{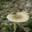} & \includegraphics[width=0.08\textwidth]{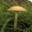} & \includegraphics[width=0.08\textwidth]{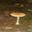} & \includegraphics[width=0.08\textwidth]{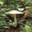}& \fcolorbox{red}{red}{\includegraphics[width=0.08\textwidth]{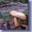}} & \includegraphics[width=0.08\textwidth]{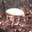}& \includegraphics[width=0.08\textwidth]{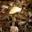} & \includegraphics[width=0.08\textwidth]{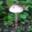}& \includegraphics[width=0.08\textwidth]{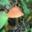} & \includegraphics[width=0.08\textwidth]{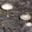} \\
    \includegraphics[width=0.08\textwidth]{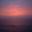} && \fcolorbox{red}{red}{\includegraphics[width=0.08\textwidth]{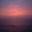}} & \includegraphics[width=0.08\textwidth]{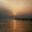} & \includegraphics[width=0.08\textwidth]{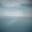} & \includegraphics[width=0.08\textwidth]{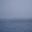}& \includegraphics[width=0.08\textwidth]{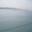} & \includegraphics[width=0.08\textwidth]{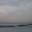}& \includegraphics[width=0.08\textwidth]{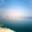} & \includegraphics[width=0.08\textwidth]{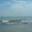}& \includegraphics[width=0.08\textwidth]{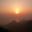} & \includegraphics[width=0.08\textwidth]{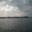} \\
    \includegraphics[width=0.08\textwidth]{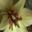} && \includegraphics[width=0.08\textwidth]{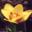} & \includegraphics[width=0.08\textwidth]{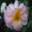} & \includegraphics[width=0.08\textwidth]{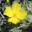} & \includegraphics[width=0.08\textwidth]{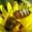}& \includegraphics[width=0.08\textwidth]{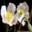} & \includegraphics[width=0.08\textwidth]{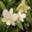}& \fcolorbox{red}{red}{\includegraphics[width=0.08\textwidth]{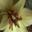}} & \includegraphics[width=0.08\textwidth]{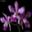}& \includegraphics[width=0.08\textwidth]{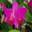} & \includegraphics[width=0.08\textwidth]{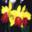} \\
    \includegraphics[width=0.08\textwidth]{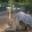} && \fcolorbox{red}{red}{\includegraphics[width=0.08\textwidth]{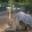}} & \includegraphics[width=0.08\textwidth]{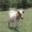} & \includegraphics[width=0.08\textwidth]{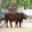} & \includegraphics[width=0.08\textwidth]{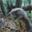}& \includegraphics[width=0.08\textwidth]{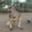} & \includegraphics[width=0.08\textwidth]{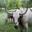}& \includegraphics[width=0.08\textwidth]{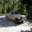} & \includegraphics[width=0.08\textwidth]{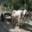}& \includegraphics[width=0.08\textwidth]{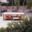} & \includegraphics[width=0.08\textwidth]{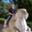} \\

    \includegraphics[width=0.08\textwidth]{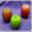} && \fcolorbox{red}{red}{\includegraphics[width=0.08\textwidth]{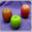}} & \includegraphics[width=0.08\textwidth]{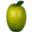} & \includegraphics[width=0.08\textwidth]{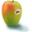} & \includegraphics[width=0.08\textwidth]{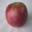}& \includegraphics[width=0.08\textwidth]{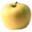} & \includegraphics[width=0.08\textwidth]{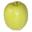}& \includegraphics[width=0.08\textwidth]{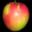} & \includegraphics[width=0.08\textwidth]{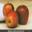}& \includegraphics[width=0.08\textwidth]{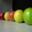} & \includegraphics[width=0.08\textwidth]{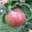} \\ \bottomrule
    \end{tabular}
    \caption{
    \textbf{Image retrieval in the quantized embedding spaces}.
    For the query image we used the quantized features of a WRN-40-2 teacher network and for the database images we used the predicted quantized features of a WRN-16-2 student network trained with our distillation method.
    As database we used the $10K$ images of CIFAR-100 test set, and as queries we used randomly sampled images from this database.
    The figure shows the query on the left-most column and top-$10$ retrieved images (in that order) next to the query. 
    We see that, as top results, we always retrieve the query itself (framed with red box) as well as other semantically and structurally similar images.
    This indicates that the two quantized embedding spaces are well aligned}
    \label{fig:retrieval}
\end{figure*}

As we claim in Section\,\ref{sec:discussion} of the main paper, 
our distillation loss enforces a quantization of the student feature maps into visual words that is in accordance to the quantization at the teacher side.
As defined in paper's Section \ref{sec:distillation_task}, we compute the soft-assignment maps $\T{\mv p}$ and $\s{\mv p}$ from $\T{\mm{f}}$ and $\s{\mm{f}}$ respectively (see equations \ref{eq:softquant} and \ref{eq:apredictor} of main paper).
In Figure~\ref{fig:retrieval}, we illustrate here the alignment of the soft-assignment maps by providing image retrieval results where the query is represented by the teacher soft-assignment map $\T{\mv p}$ while each  database image is represented by the student soft-assignment map $\s{\mv p}$.
To compute the similarity, we flatten $\T{\mv p}$ and $\s{\mv p}$ 
from $K \times H\times W$-sized tensors (where $H \times W$ are the common spatial dimensions of the teacher and student networks and $K$ is the vocabulary size) to $KHW$-dimensional vectors and then compute the dot product of the two vectors.
We see that we manage to retrieve semantically and structurally similar images, which means that the two representations $\T{\mv p}$ and $\s{\mv p}$ match well.

%% file: main.bbl
\begin{thebibliography}{10}
\providecommand{\url}[1]{\texttt{#1}}
\providecommand{\urlprefix}{URL }
\providecommand{\doi}[1]{https://doi.org/#1}

\bibitem{ahn2019variational}
Ahn, S., Hu, S.X., Damianou, A., Lawrence, N.D., Dai, Z.: Variational
  information distillation for knowledge transfer. In: Proceedings of the IEEE
  Conference on Computer Vision and Pattern Recognition (2019)

\bibitem{arandjelovic2016netvlad}
Arandjelovic, R., Gronat, P., Torii, A., Pajdla, T., Sivic, J.: {NetVLAD: CNN
  architecture for weakly supervised place recognition}. In: Proceedings of the
  IEEE Conference on Computer Vision and Pattern Recognition (2016)

\bibitem{bucilua2006model}
Buciluǎ, C., Caruana, R., Niculescu-Mizil, A.: Model compression. In:
  Proceedings of the 12th ACM SIGKDD international conference on Knowledge
  discovery and data mining. pp. 535--541. ACM (2006)

\bibitem{cho2019EKD_iccv}
Cho, J.H., Hariharan, B.: On the efficacy of knowledge distillation. In:
  Proceedings of the IEEE International Conference on Computer Vision (October
  2019)

\bibitem{csurka2004visual}
Csurka, G., Dance, C., Fan, L., Willamowski, J., Bray, C.: Visual
  categorization with bags of keypoints. In: ECCV Workshops (2004)

\bibitem{deng2009imagenet}
Deng, J., Dong, W., Socher, R., Li, L.J., Li, K., Fei-Fei, L.: Imagenet: A
  large-scale hierarchical image database. In: Proceedings of the IEEE
  Conference on Computer Vision and Pattern Recognition. pp. 248--255. Ieee
  (2009)

\bibitem{furlanello2018born}
Furlanello, T., Lipton, Z., Tschannen, M., Itti, L., Anandkumar, A.: Born-again
  neural networks. In: International Conference on Machine Learning. pp.
  1602--1611 (2018)

\bibitem{gidaris2020learning}
Gidaris, S., Bursuc, A., Komodakis, N., P{\'e}rez, P., Cord, M.: Learning
  representations by predicting bags of visual words. In: Proceedings of the
  IEEE Conference on Computer Vision and Pattern Recognition. pp. 6928--6938
  (2020)

\bibitem{girdhar2017actionvlad}
Girdhar, R., Ramanan, D., Gupta, A., Sivic, J., Russell, B.: Actionvlad:
  Learning spatio-temporal aggregation for action classification. In:
  Proceedings of the IEEE Conference on Computer Vision and Pattern
  Recognition. pp. 971--980 (2017)

\bibitem{he2016deep}
He, K., Zhang, X., Ren, S., Sun, J.: Deep residual learning for image
  recognition. In: Proceedings of the IEEE conference on computer vision and
  pattern recognition. pp. 770--778 (2016)

\bibitem{heo2019comprehensive}
Heo, B., Kim, J., Yun, S., Park, H., Kwak, N., Choi, J.Y.: A comprehensive
  overhaul of feature distillation. In: Proceedings of the IEEE International
  Conference on Computer Vision (2019)

\bibitem{heo2019knowledge}
Heo, B., Lee, M., Yun, S., Choi, J.Y.: Knowledge transfer via distillation of
  activation boundaries formed by hidden neurons. In: Proceedings of the AAAI
  Conference on Artificial Intelligence. vol.~33, pp. 3779--3787 (2019)

\bibitem{hinton2015distilling}
Hinton, G., Vinyals, O., Dean, J.: Distilling the knowledge in a neural
  network. arXiv preprint arXiv:1503.02531  (2015)

\bibitem{howard2017mobilenets}
Howard, A.G., Zhu, M., Chen, B., Kalenichenko, D., Wang, W., Weyand, T.,
  Andreetto, M., Adam, H.: Mobilenets: Efficient convolutional neural networks
  for mobile vision applications. arXiv preprint arXiv:1704.04861  (2017)

\bibitem{huang2017like}
Huang, Z., Wang, N.: Like what you like: Knowledge distill via neuron
  selectivity transfer. arXiv preprint arXiv:1707.01219  (2017)

\bibitem{jegou2010aggregating}
J{\'e}gou, H., Douze, M., Schmid, C., P{\'e}rez, P.: Aggregating local
  descriptors into a compact image representation. In: Proceedings of the IEEE
  Conference on Computer Vision and Pattern Recognition (2010)

\bibitem{JDH17}
Johnson, J., Douze, M., J{\'e}gou, H.: Billion-scale similarity search with
  {GPU}s. arXiv preprint arXiv:1702.08734  (2017)

\bibitem{kim2018paraphrasing}
Kim, J., Park, S., Kwak, N.: Paraphrasing complex network: Network compression
  via factor transfer. In: Advances in Neural Information Processing Systems.
  pp. 2760--2769 (2018)

\bibitem{krizhevsky2009learning_cifar}
Krizhevsky, A.: Learning multiple layers of features from tiny images  (2009)

\bibitem{laine2016temporal}
Laine, S., Aila, T.: Temporal ensembling for semi-supervised learning. In:
  International Conference on Learning Representations (2017)

\bibitem{li2017learning}
Li, Z., Hoiem, D.: Learning without forgetting. IEEE transactions on pattern
  analysis and machine intelligence  \textbf{40}(12),  2935--2947 (2017)

\bibitem{liu2019_cvpr}
Liu, Y., Cao, J., Li, B., Yuan, C., Hu, W., Li, Y., Duan, Y.: Knowledge
  distillation via instance relationship graph. In: Proceedings of the IEEE
  Conference on Computer Vision and Pattern Recognition (June 2019)

\bibitem{oord2018representation}
Oord, A.v.d., Li, Y., Vinyals, O.: Representation learning with contrastive
  predictive coding. arXiv preprint arXiv:1807.03748  (2018)

\bibitem{park2019RKD_cvpr}
Park, W., Kim, D., Lu, Y., Cho, M.: Relational knowledge distillation. In:
  Proceedings of the IEEE Conference on Computer Vision and Pattern Recognition
  (June 2019)

\bibitem{passalis2018learning}
Passalis, N., Tefas, A.: Learning deep representations with probabilistic
  knowledge transfer. In: Proceedings of the European Conference on Computer
  Vision. pp. 268--284 (2018)

\bibitem{peng2019CC_iccv}
Peng, B., Jin, X., Liu, J., Li, D., Wu, Y., Liu, Y., Zhou, S., Zhang, Z.:
  Correlation congruence for knowledge distillation. In: Proceedings of the
  IEEE International Conference on Computer Vision (October 2019)

\bibitem{perronnin2007fisher}
Perronnin, F., Dance, C.: Fisher kernels on visual vocabularies for image
  categorization. In: Proceedings of the IEEE Conference on Computer Vision and
  Pattern Recognition (2007)

\bibitem{quattoni2009recognizing_mit67}
Quattoni, A., Torralba, A.: Recognizing indoor scenes. In: Proceedings of the
  IEEE Conference on Computer Vision and Pattern Recognition. pp. 413--420.
  IEEE (2009)

\bibitem{romero2014fitnets}
Romero, A., Ballas, N., Kahou, S.E., Chassang, A., Gatta, C., Bengio, Y.:
  Fitnets: Hints for thin deep nets. In: International Conference on Learning
  Representations (2015), \url{https://arxiv.org/abs/1412.6550}

\bibitem{simonyan2014very}
Simonyan, K., Zisserman, A.: Very deep convolutional networks for large-scale
  image recognition. arXiv preprint arXiv:1409.1556  (2014)

\bibitem{sivic2006video}
Sivic, J., Zisserman, A.: Video google: Efficient visual search of videos. In:
  Toward category-level object recognition. Springer (2006)

\bibitem{tarvainen2017mean}
Tarvainen, A., Valpola, H.: Mean teachers are better role models:
  Weight-averaged consistency targets improve semi-supervised deep learning
  results. In: Advances in neural information processing systems. pp.
  1195--1204 (2017)

\bibitem{tian2019contrastive}
Tian, Y., Krishnan, D., Isola, P.: Contrastive representation distillation. In:
  International Conference on Learning Representations (2020)

\bibitem{tolias2013aggregate}
Tolias, G., Avrithis, Y., J{\'e}gou, H.: To aggregate or not to aggregate:
  Selective match kernels for image search. In: Proceedings of the IEEE
  International Conference on Computer Vision (2013)

\bibitem{tung2019SP_iccv}
Tung, F., Mori, G.: Similarity-preserving knowledge distillation. In:
  Proceedings of the IEEE International Conference on Computer Vision (2019)

\bibitem{yang2018knowledge}
Yang, C., Xie, L., Qiao, S., Yuille, A.: Knowledge distillation in generations:
  More tolerant teachers educate better students. arXiv preprint
  arXiv:1805.05551  (2018)

\bibitem{yim2017gift}
Yim, J., Joo, D., Bae, J., Kim, J.: A gift from knowledge distillation: Fast
  optimization, network minimization and transfer learning. In: Proceedings of
  the IEEE Conference on Computer Vision and Pattern Recognition. pp.
  4133--4141 (2017)

\bibitem{yuan2019revisit}
Yuan, L., Tay, F.E., Li, G., Wang, T., Feng, J.: Revisit knowledge
  distillation: a teacher-free framework. In: Proceedings of the IEEE
  Conference on Computer Vision and Pattern Recognition (2020)

\bibitem{zagoruyko2016WRN}
Zagoruyko, S., Komodakis, N.: Wide residual networks. In: Proceedings of the
  British Machine Vision Conference (2016)

\bibitem{zagoruyko2017AT}
Zagoruyko, S., Komodakis, N.: Paying more attention to attention: Improving the
  performance of convolutional neural networks via attention transfer. In:
  International Conference on Learning Representations (2017),
  \url{https://arxiv.org/abs/1612.03928}

\end{thebibliography}
